\documentclass[nohyperref]{article}

\usepackage{microtype}
\usepackage{graphicx}
\usepackage{subfigure}
\usepackage{booktabs} 

\usepackage{hyperref}

\usepackage[accepted]{icml2022}

\usepackage{amsmath}
\usepackage{amssymb}
\usepackage{mathtools}
\usepackage{amsthm}
\DeclareMathOperator*{\argmax}{arg\,max}

\usepackage[capitalize,noabbrev]{cleveref}

\theoremstyle{plain}

\theoremstyle{definition}

\theoremstyle{remark}

\usepackage[textsize=tiny]{todonotes}

\icmltitlerunning{Challenges and Pitfalls of Bayesian Unlearning}

\begin{document}

\twocolumn[
\icmltitle{Challenges and Pitfalls of Bayesian Unlearning}



\icmlsetsymbol{equal}{*}

\begin{icmlauthorlist}
\icmlauthor{Ambrish Rawat}{ibm}
\icmlauthor{James Requeima}{ucam}
\icmlauthor{Wessel Bruinsma}{ucam}
\icmlauthor{Richard Turner}{ucam}
\end{icmlauthorlist}

\icmlaffiliation{ibm}{IBM Research, Ireland}
\icmlaffiliation{ucam}{University of Cambridge, UK}

\icmlcorrespondingauthor{Ambrish Rawat}{ambrish@alumni.iitd.ac.in}
\icmlcorrespondingauthor{James Requiema}{jrr41@eng.cam.ac.uk}

\icmlkeywords{Machine Learning, ICML}

\vskip 0.3in
]



\printAffiliationsAndNotice{}  

\begin{abstract}

Machine unlearning refers to the task of removing a subset of training data, thereby removing its contributions to a trained model.
Approximate unlearning are one class of methods for this task which avoid the need to retrain the model from scratch on the retained data.
Bayes' rule can be used to cast approximate unlearning as an inference problem where the objective is to obtain the updated posterior by dividing out the likelihood of deleted data.
However this has its own set of challenges as one often doesn't have access to the exact posterior of the model parameters.
In this work we examine the use of the Laplace approximation and Variational Inference to obtain the updated posterior.
With a neural network trained for a regression task as the guiding example, we draw insights on the applicability of Bayesian unlearning in practical scenarios.

\end{abstract}

\section{Introduction}

Regulations like GDPR~\cite{Voigt2021_TheEU} specify a ``right to be forgotten'' which requires machine learning model providers to enable mechanisms that allow for deletion of data and/or its contributions to the learning process.
But what does it really mean for data to be deleted? In an ideal scenario deletion would result in a machine learning model that behaves identically to a model trained on the dataset that never contained the deleted data to begin with~\cite{Cao2015_Towards,Bourtoule2021_Machine,Thudi2021_Unrolling}.
In resource unconstrained settings one could arrive at such models by either retraining from scratch or maintaining checkpoints with deleted datapoints at every stage of training.
While such methods will satisfy the aforementioned deletion criterion by design, these are often prohibitively expensive to implement.
In this work we restrict our problem setting to the scenario where we no longer have access to the training data other than those to be forgotten.
We investigate some principled approaches to deletion in such settings which are motivated by a probabilistic formulation of the underlying problem.
These approaches can also be naturally used for model adoption where one may face a series of data deletion requests in conjunction with data addition~\cite{Gupta2021_Adaptive}.
Similarly, they can be used to remove erroneous data points like adversarial examples or backdoored data from a corrupted model~\cite{Liu2022_Backdoor}.

A Bayesian approach to data deletion has been previously investigated~\cite{Nguyen2020_Variational,Khan2021_Knowledge} with the work in~\citet{Nguyen2020_Variational} using the Variational Inference (VI) framework to update an approximated posterior learnt via VI with an unlearning objective. They demonstrate its usefulness for sparse Gaussian Process and logistic regression.
VI for \textit{learning} is known to be challenging to scale for Neural Networks (NN), can be costly both in terms of parameter footprint and computation time, and often suffers from issues like overconfident predictions on test data.
While a mean-field assumption helps with scaling, it can result in poor approximation \citep{Coker:2022:Wide_Mean-Field_Bayesian_Neural_Networks}.
Laplace approximation, on the other hand, is relatively inexpensive to compute and scales well for neural networks~\cite{Ritter2018_AScalable}.
In this work we adopt the Laplace approximation for the \textit{unlearning} task and contrast it with its VI counterpart.
As a case study we examine a 1-D regression task on synthetic data with removal of ``in-between'' points, similar to the work of~\citet{Foong2019_InBetween} which serves as an illustrative example of deleting informative data points.
\vspace{-2mm}
\section{Bayesian Unlearning}

Bayes' rule provides an elegant way to formulate unlearning.
As noted by~\citet{Nguyen2020_Variational}, it specifies unlearning as updating the parameter posterior by dividing the likelihood of the deleted dataset from the current posterior.
Thus, given some data for deletion, $D_\text{del}$ and the current posterior $p(\theta|D_\text{del}\cup D_\text{ret})$, the goal is to find the posterior with respect to the retained data $p(\theta|D_\text{ret})$, i.e.
\begin{align}
    p(\theta|D_\text{ret}) &\propto \frac{p(\theta|D_\text{del}\cup D_\text{ret})}{p(D_\text{del}|\theta)}.
    \label{eq:bun}
\end{align}
While in principle this formulates a simple way to update the posterior, in practice one faces a range of challenges. 
First, one usually doesn't have access to the exact posterior for non-linear models like deep neural networks due to inherent intractabilities.
The available posteriors are often computed via approximate algorithms.
Second, as we will demonstrate later, quickly decaying tails of the likelihood factor $p(D_\text{del}|\theta)$ in the denominator can destablise algorithms which try to compute or approximate $p(\theta|D_\text{ret})$.

In this work we examine approximate inference schemes for unlearning which refers to a broad class of algorithms that are used to compute parameter posteriors for a probabilistic model.
More specifically, we focus on approximate inference for Bayesian Neural Networks with  the Laplace approximation~\cite{Denker1990_Transforming, MacKay1992_APractical, Ritter2018_AScalable} and Variational Inference (VI)~\cite{Blundell2015_Weight}.
In the next section we briefly recap Laplace approximation and VI as applicable to NNs.

\subsection{Approximate Inference}
In supervised learning, you are given a dataset $D_{\text{all}} = \{(x_n,y_n)\}_{n=1}^N$, and the goal is to learn a model a $f_\theta(\cdot)$ parameterised by $\theta$, which can be used to obtain predictions $p(y^\ast|x^\ast,D_{\text{all}})$ for new data $x^\ast$.
In a Bayesian setting this is obtained by first computing a parameter posterior, $p(\theta|D_{\text{all}})$ which is subsequently used to compute the posterior predictive $p(y^\ast|x^\ast,D_{\text{all}}) = \mathbb{E}_{p(\theta|D_{\text{all}})}[p(y^\ast|x^\ast,\theta)]$.
However, exactly computing the posterior is often intractable for models like neural networks owing to their inherent non-linearities.
Therefore, one often resorts to approximations of the exact posterior, denoted $q(\theta|D_{\text{all}})$, which are learnt during training and is used for all subsequent computations ~\cite{Blundell2015_Weight}.
Numerous assumptions are made to obtain the approximate posteriors for Bayesian Neural Networks. 
For instance, the approximate posterior is often modelled as a Gaussian $q(\theta|D_{\text{all}}) = \mathcal{N}(\theta|\mu,\Sigma)$.
Furthermore, in order to scale this approach to larger models, other simplifications like independence of parameters are incorporated leading to a diagonal covariance $\Sigma$.

\textbf{Laplace Approximation.}  
Motivated as a second-order Taylor series expansion of $\log p(\theta|D_\text{all})$, the Laplace approximation formulates the approximate posterior as $\mathcal{N}(\theta|\theta_\text{MAP},\Sigma)$ where $\theta_\text{MAP} = \argmax_{\theta} (\log p(D|\theta) + \log p(\theta))$. 
With a standard Normal distribution as the prior, $p(\theta) = \mathcal{N}(\theta|0, I)$, the precision $\Sigma^{-1}$ can be computed as the negative Hessian of the likelihood at $\theta_\text{MAP}$ i.e. $I-\nabla^2_\theta \log p(D|\theta)|_{\theta_\text{MAP}}$.
In practice $\theta_\text{MAP}$ is computed via a standard Neural Network training procedure where the task loss function is augmented with a regularisation term or equivalently the gradient based optimiser is modified to include weight decay.
Computing the Hessian for a Neural Network model is a computationally expensive task and often the Gauss-Newton matrix is used as an approximation.
This only requires the computation of the Jacobian and is also guaranteed to be positive semi-definite. 
Other approximations like diagonal or blocked-diagonal assumptions or the Kronecker-Factor approximation~\cite{Ritter2018_AScalable} can further help simplify the computation and make it applicable to large scale models.
This obtained posterior can then be used to compute the predictive posterior via Monte Carlo samples~\cite{Lawrence2001_Variational} or alternatively by linearising the output of neural network around $\theta_\text{MAP}$. 
The work by \citet{Daxberger2021_Laplace} provides a comprehensive overview of Laplace approximations for Neural Network models and provides a Python library to compute the required objects.

\textbf{Variational Inference.}  
An alternate and widely used approach for approximate inference is available within the variational learning framework which aims to compute the closest distribution to the exact posterior within a family of candidate distributions $q_\psi(\theta)$ with variational parameters $\psi$. 
The closeness here is measured with the Kullback–Leibler divergence (KL) between the two distributions $\text{KL}(q_\psi(\theta)||p(\theta|D_\text{all}))$. Minimizing this KL is equivalent to maximising the Evidence Lower Bound (ELBO) to the log-marginal likelihood $\log p(D_\text{all})$ of the observed data,
\begin{align}
    \mathbb{E}_{q_\psi(\theta)} \left[\log p(D_\text{all}|\theta)\right] - \text{KL}\left(q_\psi(\theta)||p(\theta)\right).
\end{align}
In practice, the parameters $\psi$ are learnt via a Monte Carlo estimate of ELBO with stochastic optimisation which utlises the reprametrisation trick to obtain samples from $q_\psi(\theta)$.
For NNs, often an additional mean-field assumption is incorporated resulting in a fully-factored Gaussian as the chosen form for $q_\psi$~\cite{Blundell2015_Weight}.
This allows one to model a Bayesian NN with only twice as many parameters as a point-estimate model.
The predictive posterior is computed via Monte Carlo samples from the learned posterior.

The optimisation objective in both the Laplace and Variational approximations is comprised of two terms: the first term. often called the reconstruction term, includes the negative log-likelihood of the observed data which models the task objective; and the second term, be it the regularisation term in Laplace or the KL term in VI, controls the deviation from the prior.
Both these approaches present their set of pros and cons. 
While the optimisation objective presented by Laplace approximation is easy to adopt and scales well for models like NNs, it is highly localised approximating a single mode around the maximum a posteriori estimate. 
VI on the other hand presents a more challenging objective, the optimisation of which can suffer from noisy gradients. 
Moreover, it is known that the approximate posterior predictive distribution obtained from VI is often overconfident as it underestimates the variance of the exact posterior.

\subsection{Approximate Bayesian Unlearning}

Analogous to classical inference for updating the posterior after observing new data, one can unlearn the observations from an available posterior by a simple application of Bayes' rule in reverse.
In the absence of exact posterior $p(\theta|D_\text{all})$ and the retain data $D_\text{ret}$, the best one can do is use methods to obtain a distribution that approximates $1/Z\ q(\theta|D_\text{all})/p(D_\text{del}|\theta)$.
We will see that the methods we examine here balance forgetting on $D_\text{del}$ against maintaining closeness to the available approximate posterior $q(\theta|D_\text{all})$.

\subsection{L-BUN: Laplace Bayesian UNlearning}

We now formulate an analogue of the Laplace approximation for the unlearning case.
Given an approximate posterior $q(\theta|D_\text{all})$ and the $p(D_\text{del}|\theta)$ one can define the ratio of distributions $\hat{q}(\theta)$ such that,
\begin{align}
    \log \hat{q}(\theta) := -\log p(D_\text{del}|\theta) + \log q(\theta|D_\text{all}) + C.
\end{align}
One can approximate this with a second-order Taylor expansion around $\theta_\text{L-BUN} = \argmax_\theta (-\log p(D_\text{del}|\theta) + \log q(\theta|D_\text{all}))$.
Furthermore, if one were to assume that the initial approximate posterior was a Gaussian, i.e.\ $q(\theta|D_\text{all}) = \mathcal{N}(\theta|\mu_\text{all},\Sigma_\text{all})$, then $\hat{q}(\theta)$ can be approximated as a Gaussian with mean $\mu_\text{L-BUN} = \theta_\text{L-BUN}$ and precision $\Sigma^{-1}_\text{L-BUN} = \Sigma_\text{all}^{-1} + \nabla^2_\theta \log p(D_\text{del}|\theta)\Bigr|_{\theta_\text{L-BUN}}$.

The objective of this optimisation consists of two terms: the first term, $-\log p(D_\text{del}|\theta)$, encourages ``forgetting'' on the deleted dataset; and the second term, $\log q(\theta|D_\text{all})$, retains closeness to the previous posterior.
The first term aims to aggressively deteriorate performance on the deleted dataset.
In principle a model could place a likelihood of 0 at $D_\text{del}$, in which case this term is unbounded for optimisation. 
In practice one can weigh the second term with a hyperparameter $\lambda_L$ to control the optimisation dynamics.
It is worth noting that the L-BUN update bears similarity to the posterior arithmetic of EWC for continual learning~\cite{Kirkpatrick2016_Overcoming}.
Updating a model with a single step of gradient ascent along the direction of Fisher information has also been investigated for selective forgetting by~\citet{Golatkar2020_Eternal} and to unlearn backdoors by~\citet{Liu2022_Backdoor}.

\subsection{V-BUN: Variational Bayesian Unlearning}

A variational approach to Bayesian unlearning from~\citet{Nguyen2020_Variational} posits an optimisation objective to learn $\hat{q}_\psi(\theta)$ which minimises the $\text{KL}\left(\hat{q}_\psi(\theta)||\hat{p}\left(\theta\right)\right)$ where $\hat{p}(\theta) = 1/Z\ q(\theta|D_\text{all})/p(D_\text{del}|\theta))$, i.e. it is the distribution computed by dividing out the likelihood of the deleted data from the available approximate posterior.
Note that there is no assurance on how close $\hat{p}(\theta)$ is to the exact posterior $p(\theta|D_\text{ret})$ in terms of KL or otherwise, thereby limiting the interpretability of $\hat{q}_\psi(\theta)$ with respect to $p(\theta|D_\text{ret})$.
~\citet{Nguyen2020_Variational} show the equivalence of this optimisation to minimsing an Evidence Upper Bound (EUBO) to the log-marginal likelihood $\log p(D_\text{del}|D_\text{ret})$ defined as
\begin{align}
    \mathbb{E}_{\hat{q}_\psi(\theta)} \left[\log p(D_\text{del}|\theta)\right] + \text{KL}\left(\hat{q}_\psi\left(\theta\right)||p\left(\theta|D_\text{all}\right)\right).
    \label{eq:V_BUN}
\end{align}
EUBO consists of terms analogous to the L-BUN objective where the first term controls the performance on deleted data and the second term controls the deviation from the previous posterior.
~\citet{Nguyen2020_Variational} further argue that for samples of $\theta$ at the tail end of $q(\theta|D_\text{all})$ the resulting optimisation can be unstable.
They substitute $p(D_\text{del}|\theta)$ with an adjusted likelihood to stablise this optimisation which effectively ignores the gradient for samples of $\theta$ where
$q(\theta|D_\text{all}) > \lambda_V \max_{\theta^\prime} q(\theta^\prime|D_\text{all})$, i.e.\ samples which are too far from the mode of $q$.
Thus larger values of $\lambda_V$ result in dominant gradient updates from the KL term ensuring retention and smaller values enable forgetting.

\vspace{-1mm}
\section{Experiment}

\begin{figure*}[ht]
    \centering
    \label{fig:setup}
    \begin{minipage}[c]{0.19\textwidth}
    \includegraphics[width=\textwidth]{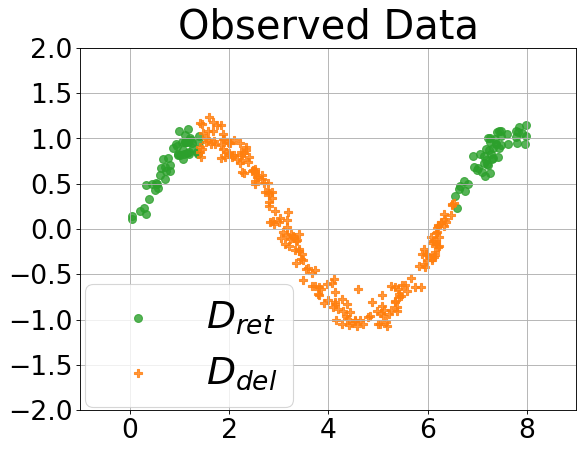}
    \end{minipage}
    \begin{minipage}[c]{0.19\textwidth}
    \includegraphics[width=\textwidth]{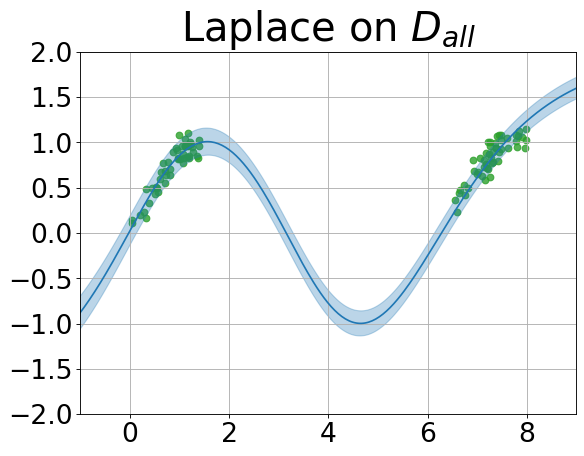}
    \end{minipage}
    \begin{minipage}[c]{0.19\textwidth}
    \includegraphics[width=\textwidth]{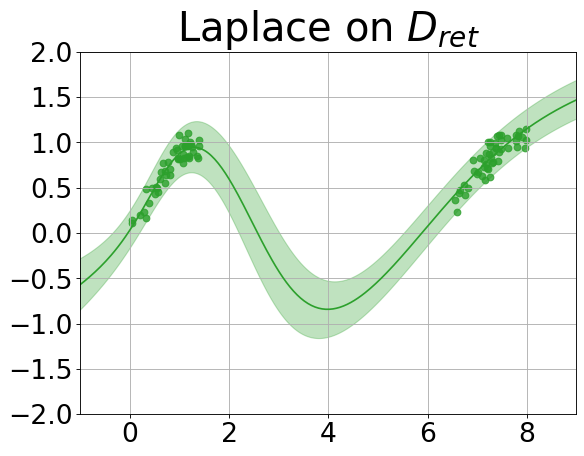}
    \end{minipage}
    \begin{minipage}[c]{0.19\textwidth}
    \includegraphics[width=\textwidth]{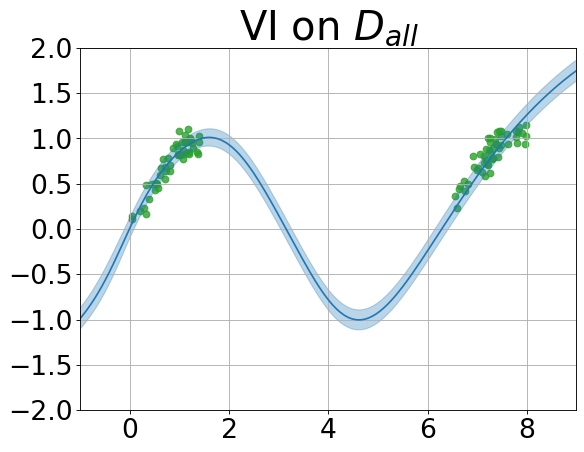}
    \end{minipage}
    \begin{minipage}[c]{0.19\textwidth}
    \includegraphics[width=\textwidth]{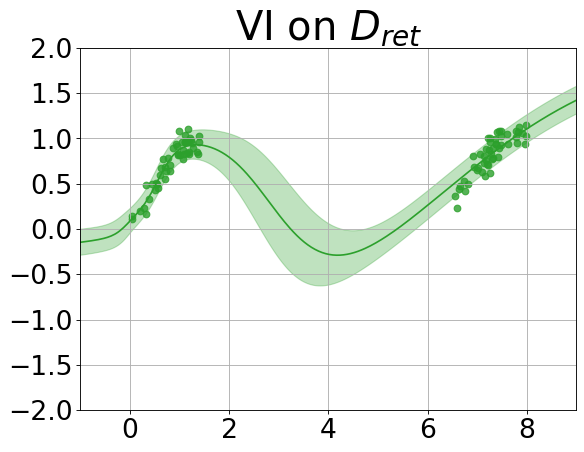}
    \end{minipage}
    \caption{Unlearning Scenario: Predictive posteriors learnt using different approximate inference schemes on all observed data $D_\text{all}$ and the retained data $D_\text{ret}$. In both cases predictions from VI are overconfident with small error bars.}
\end{figure*}

To demonstrate the various aspects of Bayesian Unlearning we examine the case of a synthetic 1D regression task where the observations $y$ are sampled from a sine wave with added Gaussian noise ($\sigma = 0.1$).
Inspired by~\citet{Foong2019_InBetween}, we consider a 1-hidden-layer Neural Network with 50 hidden units and \texttt{tanh} activation functions to model this task.
We sample 300 points which serve as the training set and observe the predictive distribution over uniformly distributed test samples for qualitative comparison.
We define the task of deletion as the removal of 200 points from the central region such that $D_\text{ret}$ comprises of two clusters of 50 input points each.
This serves as an example of deleting informative points which identifiably influences the predictive posterior, as one can observe a shifted mean with larger error bars due to lack of observed data in the deleted region. 

As a first step we obtain the posterior predictive with Laplace and VI learnt over all observed samples (Figure \ref{fig:setup}).
These serve as our starting points for unlearning.
Additionally, we also obtain the posterior predictives learnt over the retained data which serve as our \emph{target} or baselines which can be produced by retraining the model on $D_\text{ret}$ from scratch.
For all the approximations we set the observation noise to be 0.1 and model the posterior as a factored Gaussian across parameters resulting in diagonal covariance matrices.
A zero-mean prior is used across all methods. 
As shown in Figure~\ref{fig:setup}, VI is overconfident on its predictions as also noted in~\citet{Foong2019_InBetween}.

\begin{figure}[ht]
    \centering
    \label{fig:buns}
    \begin{minipage}[c]{0.22\textwidth}
    \includegraphics[width=\textwidth]{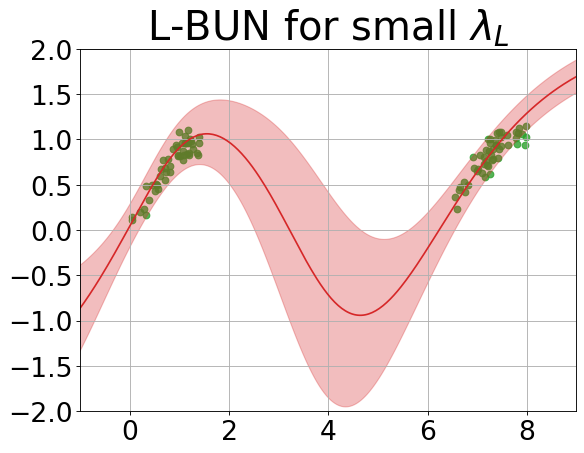}
    \end{minipage}
    \begin{minipage}[c]{0.22\textwidth}
    \includegraphics[width=\textwidth]{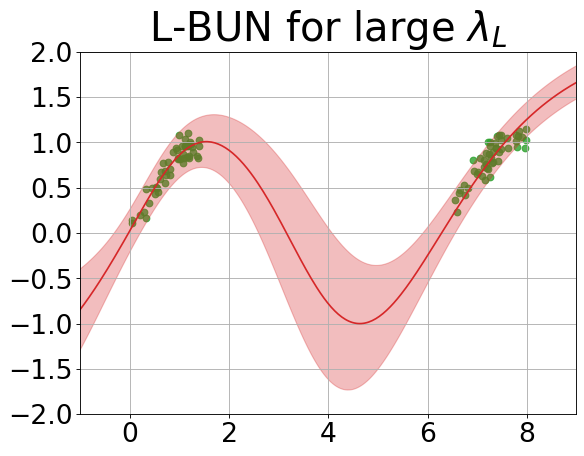}
    \end{minipage}
    \begin{minipage}[c]{0.22\textwidth}
    \includegraphics[width=\textwidth]{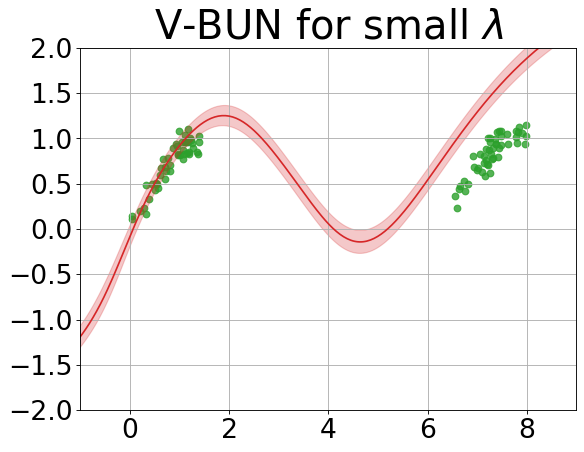}
    \end{minipage}
    \begin{minipage}[c]{0.22\textwidth}
    \includegraphics[width=\textwidth]{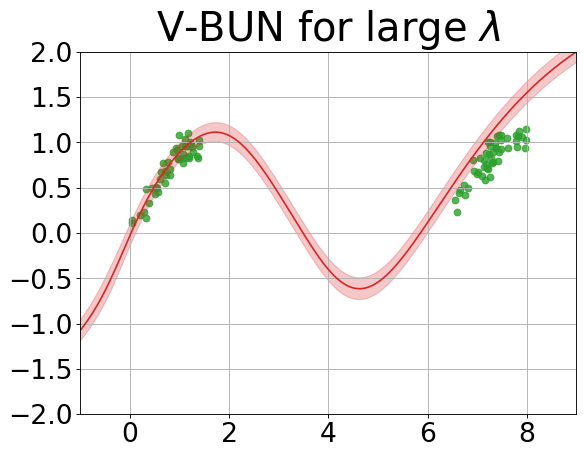}
    \end{minipage}
    \caption{Posterior predictive distributions after the deletion step for different hyperparameter settings; (Top) L-BUN is used to delete $D_\text{del}$ from the given Laplace approximation; (Bottom) VI-BUN is used for deletion from VI approximation. The shift in mean is smaller for Laplace but the uncertainties are better represented.}
\end{figure}

We use L-BUN for deletion from the Laplace approximation (top row in Fig.~\ref{fig:buns}) and V-BUN for deletion from the VI approximation (bottom row in Fig.~\ref{fig:buns}).
In both cases we use the available posteriors as the starting point for optimisation.
For L-BUN, a larger value of $\lambda_L$ ensures that the model parameters remain close to the provided posterior which we note in Figure ~\ref{fig:buns} where a smaller value of $\lambda_L$ results in a slight shift in the mean of predictive posterior and larger uncertainty in the deleted region. 
Similarly, for V-BUN a higher $\lambda_V$ results in smaller shift in predicted mean resulting from a smaller change in the posterior parameters.

\textbf{Challenges in Bayesian Unlearning.} We note that both approaches to unlearning are highly sensitive to the choice of hyperparameters ($\lambda_L$ and $\lambda_V$) with very small values leading to posteriors which place a very small (marginal) likelihood on $D_\text{del}$.
The offset from the second cluster of points in case of V-BUN (Fig.~\ref{fig:buns}) illustrates an example where one obtains a posterior that bears no similarity to the sought $p(\theta|D_\text{ret})$ or to its approximate proxies.
In the exact inference setting, dividing the posterior by the likelihood on the deleted data is guaranteed to yield a valid distribution.
However, when using an approximate posterior as our starting point, we have no such guarantees and can often run into cases where either this division or other approximations in our methods result in non-valid distributions.
If one divides out a likelihood factor that wasn't present during training, then two things can happen: (1) you end up with a distribution that doesn't normalise, so the procedure fails completely, or (2) you end up with a distribution that does normalise, but which is not the intended posterior you're after, $p(\theta|D_\text{ret})$. When using Gaussian approximations, which of the two cases you get is largely determined by the tail behaviour of the (approximate) posterior and the likelihood factor (Appendix \ref{app:path_cases}).
\citet{Luca2022_Discovering} explore a complimentary line of thought for the standard learning scenario to compare inference schemes by re-constructing the prior from a given approximate posterior.

The highly localised aspect of the Laplace approximation can exacerbate these concerns.
For instance, while the mean of a Laplace approximation is computed as a result of an optimisation step, the covariance is computed post-hoc with the local curvature.
Thus it can certainly happen that the addition and subtraction of terms from the precision matrix results in a matrix which isn't positive-semi-definite.
Similarly, given that V-BUN's target distribution is neither $p(\theta|D_\text{ret})$ not its VI approximation, its repeated application can propagate errors~\cite{Nguyen2018_Variational}.
Furthermore, use of hyperparmeter tuning approaches like use of a validation set or empirical Bayes to fit the prior~\cite{Immer2021_Scalable} can make unlearning even more challenging.

From the standpoint of unlearning optimisation, it can be seen that given enough flexibility in our model one can arrive at a model which places a likelihood of 0 on the deleted datapoints resulting in an updated posterior that can not be normalised.
Arguably, allowing the model to place a 0 likelihood is too aggressive and is not characteristic of prior assumptions on the model.
We therefore believe that information theoretic approaches like maximising the entropy of predictions on deleted data might be worth investigation.
\vspace{-2mm}
\section{Conclusion}

In this work we studied unlearning from a Bayesian perspective and contrasted two approximate inference schemes for obtaining an updated posterior of a Neural Network after data deletion.
Both these approaches cast unlearning into optimisation objectives which seek to place a small likelihood on deleted data while ensuring proximity to the original parameter posterior.
Through the lens of a synthetic example, we examined the implications of dividing a likelihood from the given approximate posterior.
It was evident that such schemes are highly sensitive to hyperparameters and can lead to scenarios which exhibit undesirable artefacts like posteriors which fail to normalise or unstable unlearning optimisation or even poor performance on retained data.
Even with this set of challenges, the simplicity of Bayesian framework and these preliminary results call for further investigation to develop methods for Bayesian unlearning.

\section*{Acknowledgements}
This work was supported by European Union’s Horizon 2020 research and innovation programme under grant number 951911 – AI4Media.

\bibliographystyle{icml2022}
\bibliography{citations.bib}

\newpage
\appendix
\onecolumn
\section{Pathological cases of Bayesian Unlearning}
\label{app:path_cases}
\begin{figure}[htp]
\centering
\includegraphics[width=.3\textwidth]{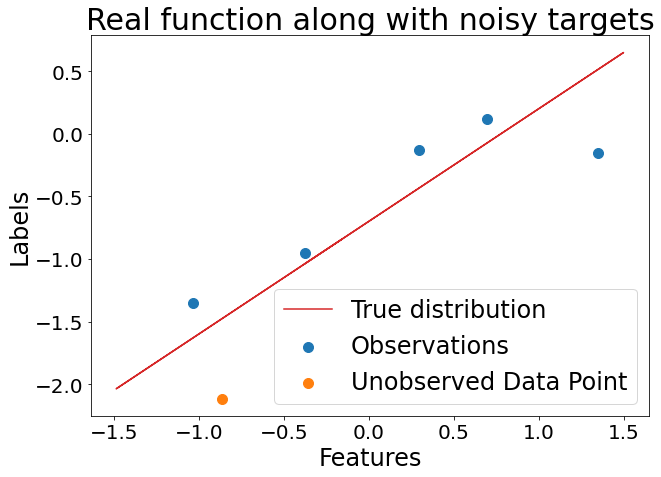}\quad
\includegraphics[width=.3\textwidth]{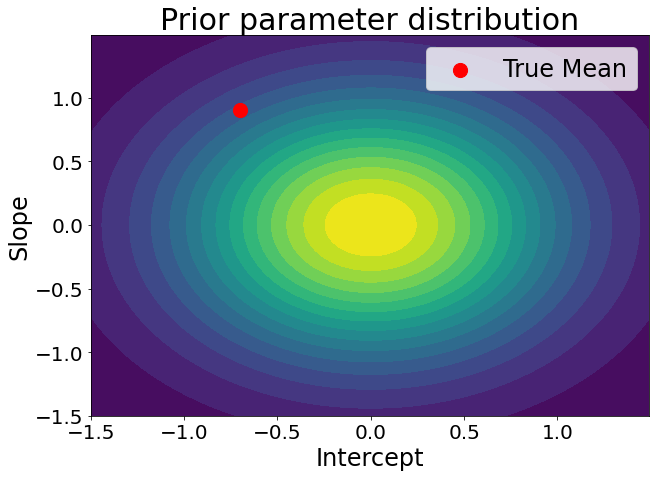}\quad
\includegraphics[width=.3\textwidth]{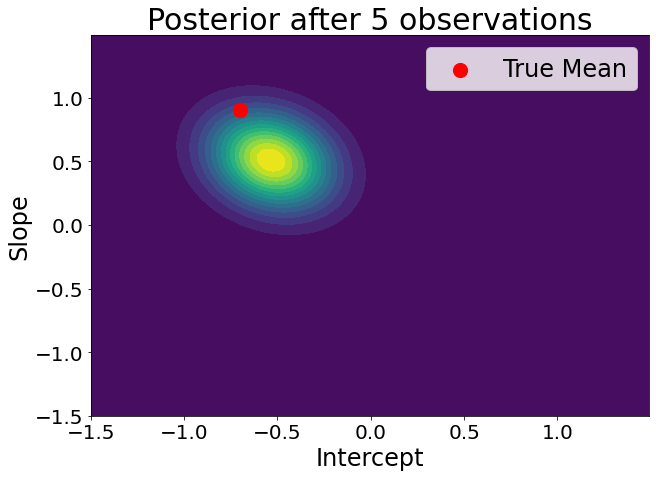}

\medskip

\includegraphics[width=.3\textwidth]{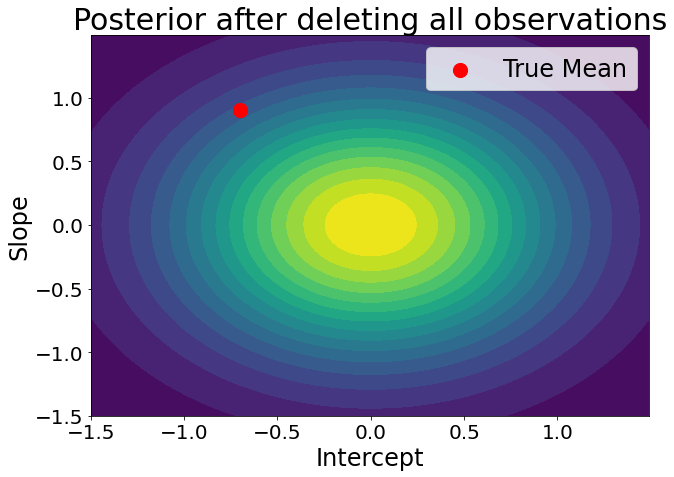}\quad
\includegraphics[width=.3\textwidth]{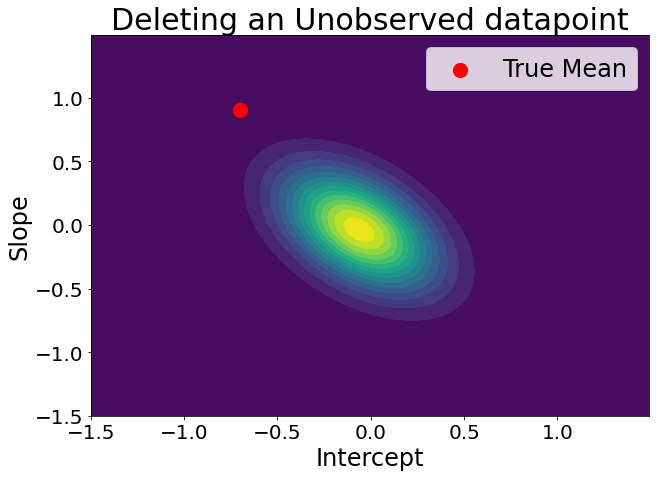}

\caption{Learning and Unlearning in Bayesian Linear Regression. (Top left) the observations (in blue), (top centre) prior parameter distribution and (top right) updated posterior for the Bayesian linear regression model after having observed 5 points. (Bottom left) the updated posterior with all 5 observations ``deteted'' which as is evident from the picture, matches the prior, followed by posterior with an unobserved datapoint (orange) ``deleted'' from the posterior. While in this case one gets a garbage posterior as shown in the picture, there are also cases when the posterior is ill-defined as the resultant covariance matrix may not be semi-positive definite.}
\label{fig:blr}
\end{figure}

To better understand the dynamics of Bayesian Unlearning, let's consider the exact case for Bayesian Linear Regression models.
Given a set of observations $(X,\mathbf{y})$, the likelihood $p(\mathbf{y}|\theta,X)$ is modelled as the Gaussian $\mathcal{N}(\boldsymbol {\phi}^T\theta,\sigma^2\mathbf{I})$ where $\boldsymbol{\phi}$ is the feature matrix for the samples $X$. Assuming a Gaussian prior $\mathcal{N}(\mu_0,\Sigma_0)$ on $\theta$, the parameter $\mu^*$ and $\Sigma^*$ for the posterior can be computed as 
\begin{align}
    \Sigma^*&=\left(\Sigma ^{-1}+ \sigma^2\boldsymbol{\phi}^T\boldsymbol{\phi}\right)^{-1}\\
    \mu^*&=\Sigma^*\left(\Sigma^{-1}\mu + \sigma^2\boldsymbol{\phi}^T \mathbf{y}\right)
\end{align}
Unlearning, as described in eq (\ref{eq:bun}) can be thought of as computing the prior parameters given the posterior parameters and data to be removed. Thus,
\begin{align}
    \Sigma&=\left({\Sigma^*}^{-1}- \sigma^2\boldsymbol{\phi}^T_{\text{del}}\boldsymbol{\phi}_{\text{del}}\right)^{-1}\\
    \mu&=\Sigma\left({\Sigma^*}^{-1}\mu^* - \sigma^2\boldsymbol{\phi}_{\text{del}}^T \mathbf{y}_\text{del}\right)
\end{align}
Figure\ref{fig:blr} shows a simple case where we begin modelling a set of observations with a linear model.
As shown, with 5 observations the mean of the posterior gets closer to the true mean.
However, the interesting case is shown in bottom-right where deleting an unobserved data point which corresponds to a likelihood term that wasn't present in the training, leads to a posterior which is uninterpretable.
In fact, if we were to remove points further away from true mean of the data generating process, deletion leads to a precision matrix that is not positive-semi-definite.

\paragraph{Tail behaviour of posterior.} We can consider similar examples to understand the tail behaviour of posterior during unlearning.
Let's consider a uni-parametric model to learn the value of a scalar after 2 observations. 
We consider
the prior
$p(\theta) \propto \exp(-\theta^4+1.5\theta^2)$ and obtain the posterior as a second-order Laplace approximation after having incorporated two likelihood factors $\exp(-(\theta+1)^2)$ and $\exp(-(\theta-1)^2)$.
This posterior is given by $1/Z\ \exp(-0.5\theta^2)$.
If we were to remove any of the two likelihood factors, the resultant function can not be normalised as the tails of $\exp(0.5\theta^2)$ blow up to infinity.

\end{document}